\documentclass[runningheads]{llncs}
\usepackage{graphicx}

\usepackage{tikz}
\usepackage{comment}
\usepackage{amsmath,amssymb} 
\usepackage{color}

\usepackage[accsupp]{axessibility}  

\usepackage{multirow}
\usepackage[ruled,linesnumbered]{algorithm2e}
\usepackage{booktabs}
\usepackage{pifont}
\usepackage{graphicx}
\usepackage{amsmath}
\usepackage{amssymb}
\usepackage{mathtools}
\usepackage{bbm}
\usepackage{wrapfig}
\begin{document}
\pagestyle{headings}
\mainmatter
\def\ECCVSubNumber{5764}  

\title{Actor-centered Representations for Action Localization in Streaming Videos} 


\titlerunning{Actor-centered representations for localization}
%
\author{Sathyanarayanan Aakur\inst{1} \and
Sudeep Sarkar\inst{2}}
\authorrunning{Aakur et al.}
%
\institute{Oklahoma State University, Stillwater, OK 74074
\email{saakur@okstate.edu}\\\and
University of South Florida, Tampa, FL, 33620
\email{sarkar@usf.edu}}
\maketitle

\begin{abstract}
Event perception tasks such as recognizing and localizing actions in streaming videos are essential for scaling to real-world application contexts. We tackle the problem of learning \textit{actor-centered} representations through the notion of \textit{continual hierarchical predictive learning} to \textit{localize} actions in streaming videos \textit{without} the need for training labels and outlines for the objects in the video. We propose a framework driven by the notion of hierarchical predictive learning to construct \textit{actor-centered} features by attention-based contextualization. The key idea is that predictable features or objects do not attract attention and hence do not contribute to the action of interest. Experiments on three benchmark datasets show that the approach can learn robust representations for localizing actions \textit{using only one epoch of training}, i.e., a single pass through the streaming video. We show that the proposed approach outperforms unsupervised and weakly supervised baselines while offering competitive performance to fully supervised approaches. Additionally, we extend the model to multi-actor settings to recognize group activities while localizing the multiple, plausible actors. We also show that it generalizes to out-of-domain data with limited performance degradation. 
\end{abstract}

\section{Introduction}\label{sec:intro}
\begin{figure}[t]
    \centering
    \includegraphics[width=0.99\columnwidth]{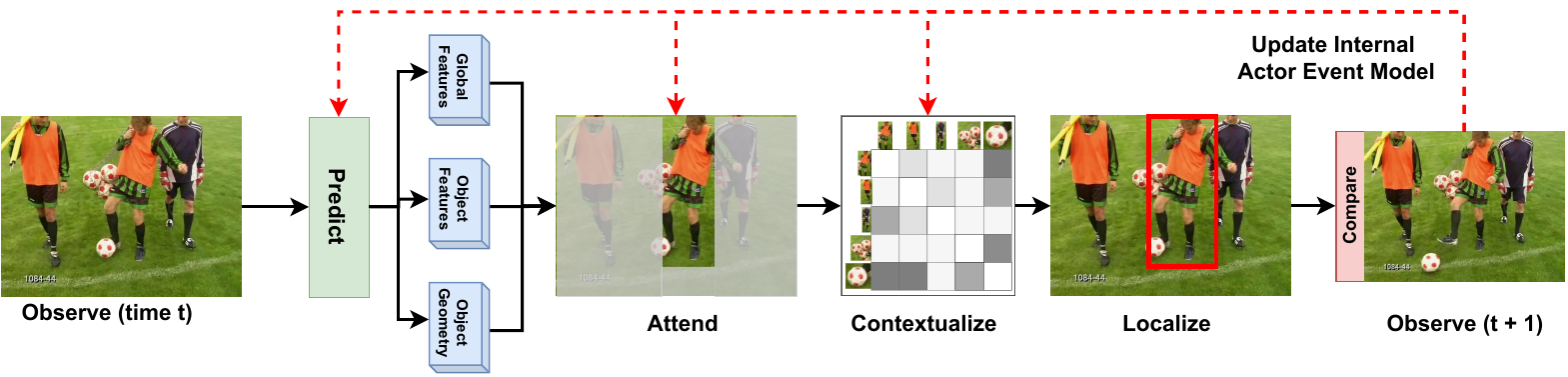}
    \setlength{\belowcaptionskip}{-15pt}
    \caption{Our goal is to learn actor-centered representations for actor localization in streaming videos \textit{without explicit annotations}. Given a frame at time $t$, we follow the sequence of: observe, predict, attend based on prediction error, contextualize actor representations, and localize. The internal event models are constantly updated based on the prediction errors with the observation at time $t+1$.}
    \label{fig:intuition}
\end{figure}
Understanding events in videos requires understanding beyond recognition, such as localizing the actor, understanding their future behavior from current and past observations, and building robust representations at the event and actor levels. While many recent works have focused on action recognition~\cite{aakur2019wacv,kuehne2014language,li2018videolstm} and action localization~\cite{escorcia2020guess,li2021groupformer,gavrilyuk2020actor}, significant progress has primarily been driven by the use of large-scale, annotated training data. 
While self-supervised learning ~\cite{gan2018geometry,wang2019self} has reduced the need for labeled data for \textit{recognition}, there is still a dependency on large amounts of manual annotations for \textit{localization}.

We consider the problem of learning \textit{actor-centered} representations to localize actions in \textit{streaming videos} i.e., needing a single-pass through video for training (single epoch) and without training labels and outlines. We do not need multiple training epochs to build the representations.
We define an actor-centered representation as a compositional structure of the scene that encodes the properties (location, geometry, and relational cues) of the \textit{dominant} actor contributing to the action of interest. For example, in Figure~\ref{fig:intuition}, there are many actors (three players, a soccer ball, etc.) in the scene, but only one \textit{dominant actor} (the player in the middle) is involved in the action ``\textit{kicking ball}''. Hence, an actor-centered representation would encode the appearance and geometry of the player in the middle and \textit{contextualize} their features concerning the other objects in the scene. 
Such representations allow us to capture action-specific contextual cues in a generalizable representation. 

We build actor-centered event representations by contextualizing the actor's features with environment-level (or scene) at both a perceptual level (such as color, texture, and movement) and a conceptual level (such as actor-environment interactions and action goals). 
Computationally, we model this process by following a sequence of operations given by \textit{observe, predict, compare, attend, contextualize}, and \textit{localize}. Figure~\ref{fig:intuition} illustrates this process. The \textit{key idea} is that predictable features or objects do not attract attention and hence do not contribute to the action of interest.
We introduce the idea of hierarchical prediction that enables the framework to select objects of interest and maintain context in prediction to localize the action by navigating spurious motion patterns such as camera motion and background clutter. 
This hierarchical prediction differs from prior versions of predictive learning for action recognition~\cite{Aakur_2019_CVPR,aakur2020action}, which do not consider the actor-centered features such as appearance, geometry, and their evolution with respect to the scene. 

The \textbf{contributions} are four-fold: (i) we introduce the idea of \textit{hierarchical} predictive learning to learn actor-centered representations to localize actions in \textit{streaming videos} in an \textit{unsupervised manner}, (ii) introduce a novel, attention-driven formulation for learning robust, actor-centered event features for action localization and recognition, (iii) demonstrate that the proposed approach can be trivially extended to multi-actor group activity recognition and localization, and (iv) show that the use of actor-centered feature representations helps learn robust features that can generalize to data from outside the training domain \textit{without finetuning} \textit{for both localization and group activity recognition}.

\section{Related Work}\label{sec:related}
\textbf{Action localization} has largely been tackled through \textit{supervised} learning approaches~\cite{gkioxari2015finding,hou2017tube,jain2017tubelets,soomro2015action,soomro2016predicting,tian2013spatiotemporal,tran2012max,wang2014video,weinzaepfel2015learning}, which aim to simultaneously generate bounding box proposals and labels learned from annotated training data. The common pipeline uses convolutional neural networks (both 2D and 3D~\cite{tran2015learning}) to extract features from RGB images, optionally the optical flow images, and generate bounding box proposals to localize objects in the video sequence. A linking algorithm (Viterbi or actor linking~\cite{escorcia2020guess}) is used to extract action tubes from the generated bounding boxes. Annotated training data is used to train recognition and bounding box regression modules. 

\textit{Weakly supervised}~\cite{escorcia2020guess,li2018videolstm,sharma2015action} reduce the dependency on training data by negating the need for spatial-temporal annotations and using either attention-based pooling~\cite{li2018videolstm,sharma2015action} or appearance-based linking from generic object detection-based proposals~\cite{escorcia2020guess}. They typically require video-level label annotations that are used to learn representations for recognition and use object-level labels and characteristics to select bounding box proposals from pre-trained object detection models. Hence, they may be constrained to localizing actions specific to classes from the detection models.

\textit{Unsupervised} approaches~\cite{aakur2020action,soomro2017unsupervised} do not require annotations for labels or bounding boxes. Soomro \textit{et al.}~\cite{soomro2017unsupervised} use pre-trained object detection models to generate proposals and score each with a ``humanness'' score that ranks the likelihood of belonging to an action class and uses a knapsack-based algorithm to discover action classes to self-label videos. Aakur \textit{et al}~\cite{aakur2020action} use a predictive learning-based approach (PredLearn for brevity) to create spatial-temporal attention maps which are used to localize objects of interest. Closely related to our approach, PredLearn anticipates the future spatial feature using a motion-weighted loss function at the feature level. However, it does not enforce consistency in actor-specific features such as geometry or contextualized representations to help reject the background clutter and maintain context in prediction. Additionally, we localize multiple actors together with learning robust features.

\section{Actor-centered Action Localization}\label{sec:proposed}
\begin{figure*}[t]
    \centering
    \includegraphics[width=0.95\textwidth]{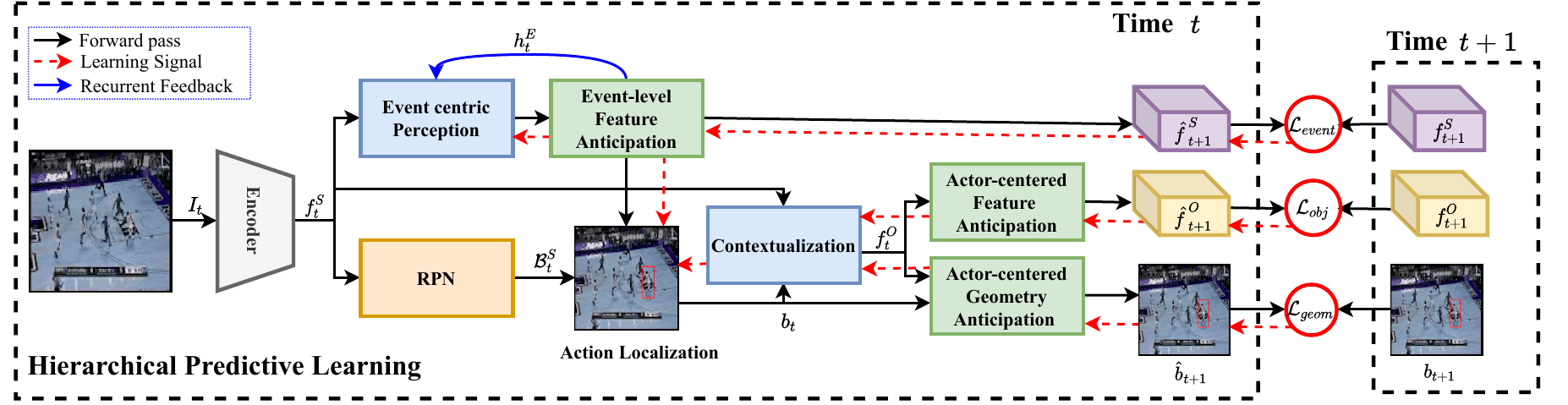}
    \setlength{\belowcaptionskip}{-15pt}
    \caption{\textbf{Overview of our approach.} Given a sequence of frames in \textit{streaming fashion}, our model constructs an actor-centered representation using the notion of hierarchical predictive learning. A prediction-driven attention map is used to localize the action.}
    \label{fig:overall_arch}
\end{figure*}

\textbf{Problem Formulation.} In our setup, we consider the problem of localizing the \textit{dominant} action $a_i$ at each time instant $t$ in a \textit{streaming video}. Each video can contain multiple objects in the scene with one \textit{dominant} action performed by one or more \textit{actors}. The key challenge is ignore clutter and identify the object(s) of interest (i.e., the \textit{actors}) \textit{without any supervision} while building robust representations that capture the motion and relational dynamics of the event. 
Figure~\ref{fig:overall_arch} illustrates the proposed action localization framework. 
We begin with perceptual features extracted from a convolutional neural network~\cite{simonyan2014very} and progressively refine these scene-level features with context from event-level dynamics (Section~\ref{sec:event-centric-percep}) and actor-centered context (Section~\ref{sec:contextualization}) through the notion of hierarchical predictive learning (Section~\ref{sec:hier_pred}), to jointly model both the evolution of the action and the actors in a unified framework. 

\subsection{Extracting Perceptual Features}\label{sec:spatial}
First, we extract a global, scene-level representation of the given visual sequence. This representation includes both perceptual features and identifying regions of likely interest representing objects. While our approach is general enough to handle different object proposal approaches (see Section~\ref{sec:results}), we use a pre-trained convolutional neural network to extract the scene-level representation ($f^{S}_t$) and use a Single-Short Object Detector (SSD)~\cite{liu2016ssd} layer to generate region-proposals ($\mathcal{B}^S_t$), where objects are likely to exist. Following prior work in \cite{aakur2020action}, we make the SSD class-agnostic by considering \textit{all} bounding boxes returned (at an ``objectness'' threshold of $0.01$) regardless of the predicted class and their corresponding confidence scores. This allows us to remove biases towards certain actors, such as human actors, and help handle any visual variations (such as pose and occlusion) that can cause missed detection. 

\subsection{Event-centric Perception}\label{sec:event-centric-percep}
The second step in the proposed model is to construct a feature representation of the current scene (at time $t$) influenced by the observed event's spatial-temporal dynamics. While CNN features provide an efficient spatial representation, it does not consider the contextual knowledge provided by temporal transitions and spatial interactions among the scene's entities. This process requires modeling a stable event representation $h^E_t$ and an attention mechanism $\alpha^S_t$ that uses this global representation to jointly perceive and anticipate the spatial and temporal dynamics of the event. 
Formally, we define the event-centric perception model as a prediction function that maximizes the probability $P(\hat{f}^S_{t+1}|W_p, \alpha^S_t, h^E_t, f^S_t)$, where $\hat{f}^S_{t+1}$ is the anticipated features at time $t+1$ conditioned on an internal event representation $h^E_t$, a temporally-weighted, spatial attention function $\alpha^S_t$ and the current observed features $f^S_t$. $W_p$ is the set of learnable parameters in this module. 
There are two steps in constructing the event-centric perceptual features - (i) learning an efficient global, event-level representation and (ii) using the learned event representation to drive the perception in a recurrent manner. 
First, we first create the event-centric scene representation by weighting the CNN feature $f^S_t$ of the frame at time $t$ by an attention vector ($\alpha^S_t$), influenced by the spatial temporal dynamics of the current event. Hence, the event-centric representation is given by $f^E_t  {=} \alpha^S_t \odot f^S_t$, where $\alpha^S_t {=} f_a(f^S_t, h^E_{t-1})$ and $f_a(\cdot)$ is a learned attention function~\cite{bahdanau2014neural}. 
Second, we use a hierarchical stack of Long Short Term Memory (LSTM) networks \cite{hochreiter1997long} to construct the internal event representation. We take a continual predictive learning approach, inspired by \cite{Aakur_2019_CVPR,aakur2020action}, to learn an efficient global representation of the event that captures the relevant spatial-temporal patterns to provide context for event-based perception. 
The hierarchical LSTM stack is used as a spatial-temporal decoder network. It takes a sequence of event-centric image features as input and propagates its prediction up the stack. The output of the top-most LSTM is taken as the anticipated features ($\hat{f}^S_{t+1}$) at the next time step. Hence, the hierarchical LSTM stack acts as a generative model that learns and uses a stable event representation to anticipate the scene's spatial and temporal evolution. 
Formally, this is represented as 
\begin{align}
     \hat{f}^\ell_{t+1}, h^{{\ell}}_t = LSTM(\hat{f}^{\ell-1}_{t+1}, W_\ell, h^{{\ell}}_{t-1})\label{eqn:predLayer}\\
     \hat{f}^0_{t+1}, h^{{0}}_t = LSTM(f^E_t, W_0, h^{{0}}_{t-1})\label{eqn:pred0}
\end{align}
where $\hat{f}^{\ell-1}_{t+1}$ refers to the predicted features at the $\ell^{th}$ LSTM in the stack and $W^\ell$ refers to the weights associated with the LSTM at the $\ell^{th}$ layer; Equation~\ref{eqn:pred0} shows the initialization for the bottom-most $0^{th}$-level LSTM.

Note that $\hat{f}^\ell_{t+1}$ for the top most LSTM network is taken as the prediction for time $t+1$ and the corresponding hidden state is taken as the event representation such that $h^E_t{=}h^{{\ell}}_t$. The memory is not shared within the stack and hence allows each level of the stack to model the spatial-temporal dynamics at different granularity, with the $\ell^{th}$-level LSTM influenced by the lower-level LSTMs. 

\textbf{Event-level prediction.} The event-centric perception module is trained in a predictive learning approach with the training objective given by
\begin{equation}
    \mathcal{L}_{event} = {\lVert f^S_{t+1} - {f}^S_{t}\lVert_{2}} \odot {\lVert f^S_{t+1} - \hat{f}^S_{t+1}\lVert_{2}} 
    \label{eqn:globalPred}
\end{equation}
where the first term represents the weighted difference between the \textit{features} at consecutive time steps $t$ and $t+1$ and the resulting value $\mathcal{L}_{event}$ represents a weighted $L-2$ norm of the predicted and expected value that penalizes incorrect predictions at spatial locations with maximal change at the \textit{feature level}. Hence, $\mathcal{L}_{event}$ is the prediction-based drive, a measure of the effectiveness of the learned event representation at a coarse spatial quantization. This predictive learning process forms the bottom level of the hierarchy and helps learn event-level dynamics for modeling the action in the scene. While effective, as shown in PredLearn~\cite{aakur2020action}, it is not enough to handle complex scenes and multiple actors in the scene. There is a need to model the actor-environment interactions for effective visual understanding. 

\subsection{Contextualization: Actor-centered Features}\label{sec:contextualization}
The next step is to construct actor-centered representations that contextualize the event-level dynamics with the actor-environment interactions. We consider a feature representation of a scene to be \textit{actor-centered} if the resulting representation can (i) reject clutter in the scene, (ii) reduce the impact of background or spurious motion patterns, and (iii) \textit{contextualize} the actor's motion dynamics with the rest of the scene or environment. This representation is analogous to a posterior-weighted spatial representation that highlights areas of interest while suppressing spatially irrelevant features. In our framework, the posterior is obtained by updating the prior, captured by the prediction-based error from Equation~\ref{eqn:globalPred}, with the current observation ($f^{S}_t$). The contextualized representation is obtained by computing the dot-product attention~\cite{luong2015effective} between the posterior-weighted representation and the actual representation and is defined as
\begin{equation}
   f^{O}_t = GAP(softmax({f^{S}_t\odot\mathcal{F}^S_t})\odot \mathcal{F}^S_t) \label{eqn:object_contextualization}
\end{equation}
where $GAP$ refers to the Global Average Pooling function~\cite{lin2013network} and $\mathcal{F}^S_t$ is a contextualized feature representation conditioned by the posterior probability provided by the spatial-temporal prediction loss $\mathcal{L}_{Event}$. We compute this function as $\mathcal{F}^S_t {=} softmax(\mathcal{L}_{Event})\odot f^{S}_t$, which intuitively provides a representation that rejects clutter by scaling down the spatial regions that do not contribute to the prediction uncertainty. These areas typically involve background scenes or actors whose actions are more predictable and less likely to be of interest. 
This formulation helps preserve the scene's spatial-temporal structure by summing out any trivial motion-based changes, making it more robust to spurious motion patterns in the input, such as those induced by background noise and small camera motion. 
Empirically, this formulation results in a more robust video-level representation that can generalize across domains (see Section~\ref{sec:quant_results}).

\subsection{Hierarchical Predictive Learning}\label{sec:hier_pred}
We use the notion of \textit{continual, hierarchical predictive} learning to train the model end-to-end without needing labels and outlines of the objects in the video. This approach aims to model the dynamics of the observed event at different levels of granularity, moving beyond just scene-level dynamics \cite{aakur2020action} or temporal dynamics~\cite{Aakur_2019_CVPR}. 
To this end, we create a hierarchy of predictions that are performed at every time step that models the event-level and actor-level dynamics in the event. At the lowest level is the event-level prediction (Section~\ref{sec:event-centric-percep}). At the next level is the prediction of the actor's dynamics within the scene's context. At the top level is the prediction of the actor's visual properties. The goal is to anticipate the actor's location and geometry in the context of event-level and actor-level dynamics. Each level of the stack influences the prediction of the upper level and hence forms a hierarchy of predictions that capture the inherent dynamics within the event. The actor-level predictions (levels 2 and 3) are conditioned on the event-level representation by constructing a global representation given by $\hat{f}^E_{t} = GAP(\alpha^S_t\odot \hat{f}^S_{t})$, where $\hat{f}^S_{t}$ refers to the anticipated spatial features (from the previous prediction step) and $\alpha^S_t$ refers to the spatial attention constructed (conditioned on the current observation) at time $t$. This formulation of the global representation forces the model to learn spatially relevant features that are important \textit{across time steps} and hence helps ensure that the event representation is robust by acting as temporal smoothing. 

Computationally, we learn two LSTM-based prediction models that use this global representation ($\hat{f}^E_{t}$) to anticipate the actors' dynamics in terms of contextualized features and geometry. One LSTM anticipates the \textit{changes} in the actor's geometry rather than directly predicting the BB location, which allows the predictor to focus on the \textit{evolution} of geometry. The other LSTM anticipates the actor-centered representations ($\hat{f}^O_{t+1}$) at time $t+1$. Hence, the goal of these two LSTMs is to minimize the actor-centered prediction errors defined as 
\begin{equation}
    \mathcal{L}_{object} = {\lVert f^O_{t+1} - \hat{f}^O_{t+1}\lVert_{2}} +  \mathcal{D}_{bb}(b_{t+1}, \hat{b}_{t+1}) + \mathcal{D}_{g}(b_{t+1}, \hat{b}_{t+1})
    \label{eqn:objectPred}
\end{equation}
where $\mathcal{D}_{bb}(b_{t+1}, \hat{b}_{t+1})$ is the distance between the predicted bounding box and the actual observed bounding centers; $\mathcal{D}_{g}(b_{t+1}, \hat{b}_{t+1}){=}(\sqrt{w}-\sqrt{\hat{w}})^2 +(\sqrt{h}-\sqrt{\hat{h}})^2$, where ($\hat{h}$ and $\hat{w}$) and (${h}$, ${w}$) are the predicted and actual height and widths of the bounding box $bb_{t+1}$, respectively. 
Hence, the entire framework is trained end-to-end using the overall objective function given by
\begin{equation}
    \mathcal{L}_{total} = \lambda_{1}\frac{1}{n_f}\sum^{w_f}_{i=1}\sum^{h_f}_{j=1} \mathcal{L}_{event} + \lambda_{2}\mathcal{L}_{object} 
\end{equation}
where $\lambda_1$ and $\lambda_2$ are modulating factors to balance the trade-off between predicting the event-level and object-level prediction errors. Both losses directly penalize the event-level representation ($h^E_t$), the spatial attention ($\alpha^S_t$) and the contextualization module ($\mathcal{F}^S_t$. Hence it adds an implicit regularization to prevent overfitting since the model's parameters are updated \textit{continuously per frame}. The resulting spatial-temporal loss $\mathcal{L}_{event}$ can then considered to be reflective of the \textit{predictability} of both the actor \textit{and} scene. Hence, spatial locations with a higher error indicate the location of the actor~\cite{aakur2020action,horstmann2015surprise}. Note that the entire process is unsupervised, there are no labels or bounding box annotations needed for training since the predictions at time $t$ are compared to observations at time $t+1$ to provide supervision to progressively refine the representations. 

\subsection{Attention-based Action Localization}\label{sec:localization}
The final step in the proposed approach is using attention to localize the actor (the object of interest) in the given video. We create an attention-like representation using the prediction-based error $\mathcal{L}_{event}$ to identify areas of interest. 
The input to the localization process consists of (i) initial regions of interests generated based on spatial features $\mathcal{B}^S_t$ (from Section~\ref{sec:spatial}), (ii) the spatial-temporal prediction error $\mathcal{L}_{event}$ (from Section~\ref{sec:event-centric-percep}), (iii) number of attention ``grids'' to consider $K$, and (iv) the total number of bounding box predictions per frame $t$. We first construct an attention-like representation by running the spatial-temporal prediction error through a \textit{softmax} function to produce an attention map of shape $c_x\times c_y$ where $c_x$ and $c_y$ are spatial dimensions of the observed feature maps, with each point corresponding to a ``grid'' in the frame (following notation from YOLO~\cite{redmon2017yolo9000}). The softmax operation magnifies areas of high errors while suppressing areas of low prediction errors. 

We consider areas of high prediction error to be regions of interest. However, we allow the attention map to be split between multiple objects and consider the top $K$ grids (sorted based on prediction error) to select bounding box localization. Following the notation from YOLO-based object detection models~\cite{redmon2017yolo9000}, we define a binary function $\mathbbm{1}(\cdot)$ that returns \textit{True} if a bounding box proposal's center falls within the ``grid'' $e_{i,j}$ and \textit{False} otherwise. This allows us to select objects that are most likely to contribute to the grid's prediction error. Note that this is different from \cite{aakur2020action}, where each bounding box is assigned an energy term based on distance from a prior position and the magnitude of the prediction error, which does not allow them to attend to multiple objects. 
This is further explored in Section~\ref{sec:results}, where the use of hierarchical prediction allows the model to attend to multiple objects simultaneously for multi-actor localization. 

\textbf{Implementation Details.} In our experiments, we use a VGG-16 network~\cite{simonyan2014very}, pre-trained on ImageNet~\cite{ILSVRC15}, as the backbone network for training a Single Shot Multbox Detector (SSD)~\cite{liu2016ssd} to extract frame-level representations and generate localization proposals. The SSD is trained on MS-COCO with input re-sizes to $512\times 512$. We use the output of the max-pooling layer after the fifth convolutional layer as $f^S_t$. We use the SSD as a class-agnostic region proposal network by taking the bounding box proposals without any predicted classes or associated probabilities. The number of layers $\ell$ in the hierarchical prediction network (in Section~\ref{sec:event-centric-percep}) as $3$ and set the dimensions of the hidden state at each layer to $512$. 
We set the number of attention grids $K=5$ and the number of localization per frame $N=10$ (Section~\ref{sec:localization}). 
We train with adaptive learning~\cite{Aakur_2019_CVPR}, with an initial learning rate of $1\times10^{-10}$ and scaling factors $\Delta^{-}_{t}=0.1$ and $\Delta^{+}_{t}=0.01$. 
\section{Experimental Evaluation}\label{sec:results}
\subsection{Data, Metrics and Baselines}
We use three standard benchmark datasets (UCF Sports~\cite{rodriguez2008action}, JHMDB~\cite{jhuang2013towards}, and THUMOS'13\cite{jiang2014thumos}) to evaluate the proposed approach for action localization. We also evaluate on the Collective Activity dataset~\cite{choi2009they} to demonstrate and evaluate our approach on multi-actor action localization.
\textbf{UCF Sports}~\cite{rodriguez2008action} contains $10$ classes characterizing sports-based actions such as weight-lifting and diving. We use the official splits containing $103$ videos for training and $47$ videos for testing, as defined in ~\cite{lan2011discriminative} for evaluation. 
\textbf{JHMDB}~\cite{jhuang2013towards} has $21$ action classes from $928$ trimmed videos, each annotated with human joints and bounding box for every frame. 
It offers several significant challenges for unsupervised action localization, such as camera motion that causes significant occlusions and background objects that act as distractions. We report all results as the average across all three splits. 
\textbf{THUMOS'13}~\cite{jiang2014thumos} (or the \textit{UCF-101-24} dataset) is a subset of the UCF-101~\cite{soomro2012ucf101} dataset, consisting of $24$ classes and $3,207$ videos. It is one of the most challenging action localization datasets with complex motion, background clutter and high intra-class variability. Following prior works~\cite{li2018videolstm,soomro2017unsupervised}, we report results on the first split. 
\textbf{Collective Activities}~\cite{choi2009they} is a group activity dataset where the goal is to recognize the activity performed by multiple actors such as talking, queueing, and walking. It comprises 44 short
video sequences with 5 group activities, with every 10 frames annotated with bounding boxes of actors involved in the group activity. This dataset offers a unique challenge in localizing all actors ($>=1$) involved in the action while learning robust features that can capture the dynamics of each actor in the context of their collective activity. We follow prior works~\cite{wu2019learning,ibrahim2016hierarchical,wang2017recurrent,qi2018stagnet,gavrilyuk2020actor} and use $1/3$ of the video sequences for testing and the rest for training. We report results for both recognition and localization. 

\begin{table*}[t]
\centering
\begin{tabular}{|c|c|c|c|c|c|c|c|c|}
\toprule
\multirow{2}{*}{\textbf{Approach}} & \multicolumn{2}{|c|}{\textbf{Supervision}}  &  \multicolumn{2}{|c|}{\textbf{UCF Sports}}  &  \multicolumn{2}{|c|}{\textbf{JHMDB}}  &  \multicolumn{2}{|c|}{\textbf{THUMOS'13}}  \\
\cline{2-9}
 & \textbf{Spatial} & \textbf{Label} & $\sigma{=}$\textbf{0.2}  & $\sigma{=}$\textbf{0.5}  & $\sigma{=}$\textbf{0.2}  & $\sigma{=}$\textbf{0.5} & $\sigma{=}$\textbf{0.2}  & $\sigma{=}$\textbf{0.5} \\
 \toprule
 Tube CNN~\cite{hou2017tube} & \ding{51} & \ding{51} & 0.47 & - & - & {0.77} & 0.47 & 0.41\\
 Action Tubelets~\cite{jain2017tubelets} & \ding{51} & \ding{51} & 0.53 & 0.27 & - & - & {0.48} & - \\
 Action Tubes~\cite{gkioxari2015finding} & \ding{51} & \ding{51} & {0.56} & {0.49} & 0.55 & 0.45 & - & - \\
 MRSTL~\cite{zhang2020learning} & \ding{51} & \ding{51} & - & - & - & 0.37 & - & 0.68 \\
 MENET~\cite{liu2020real} & \ding{51} & \ding{51} & - & - & - & \textbf{0.82} & - & \textbf{0.84} \\
 HISAN~\cite{pramono2019hierarchical} & \ding{51} & \ding{51} & - & - & - & 0.77 & - & 0.73 \\
 ACAR-Net~\cite{pan2021actor} & \ding{51} & \ding{51} & - & - & - & - & - & 0.84 \\
 \midrule
 ALSTM~\cite{sharma2015action} & \ding{55} & \ding{51} & - & - & - & - & 0.06 & - \\
 VideoLSTM~\cite{li2018videolstm} & \ding{55} & \ding{51} & - & - & - & - & 0.37 & - \\
 Actor Supervision~\cite{escorcia2020guess} & \ding{55} & \ding{51} & - & \textbf{0.48} & - & \textbf{0.36} & \textbf{0.46} & - \\
 \midrule
 Soomro \textit{et al}~\cite{soomro2017unsupervised} & \ding{55} & \ding{55} & 0.46$^*$ & 0.30$^*$ & \textbf{0.43}$^*$ & \textbf{0.22}$^*$ & 0.21$^*$ & 0.06$^*$\\
 PredLearn~\cite{aakur2020action} ($k{=}k_{gt}$) & \ding{55} & \ding{55} & 0.55 & 0.32 & 0.30 & 0.10 & 0.31 & 0.10\\
AC-HPL (Ours, $k{=}k_{gt}$) & \ding{55} & \ding{55} & \textbf{0.70} & \textbf{0.59} & \textbf{0.43} & 0.15 & \textbf{0.38} & \textbf{0.20}  \\
 \bottomrule
\end{tabular}
\setlength{\belowcaptionskip}{-15pt}
\caption{Comparison with state-of-the-art approaches on three common benchmark datasets - UCF Sports, JHMDB and THUMOS'13. We report the video-level mAP at different overlap thresholds. ${^*}$ refers to the use of class-specific object proposals. }
\label{tab:sota}
\end{table*}

\textbf{Label Prediction and Metrics.} Due to the unsupervised nature of learning representations, we use \textit{k-means} clustering to obtain class labels. The frame-level features are max-pooled to obtain video-level features. Following prior work~\cite{aakur2020action,ji2019invariant,xie2016unsupervised}, we use the Hungarian method to map from predicted clusters to the ground-truth labels. We set the number of clusters to the number of classes in the ground-truth for comparison with state-of-the-art. 
For \textit{action localization}, we report the mean average precision (mAP) metric at different overlap thresholds for a fair comparison with prior works~\cite{li2018videolstm,soomro2017unsupervised}. 

\textbf{Baselines.} We compare against several fully supervised baselines such as MRSTL~\cite{zhang2020learning}, MENET~\cite{liu2020real}, HISAN~\cite{pramono2019hierarchical}, ACAR-Net~\cite{pan2021actor}, tube convolution networks~\cite{hou2017tube}, motion-based action tublets~\cite{jain2017tubelets} and action tubes~\cite{gkioxari2015finding} and weakly supervised such as ALSTM~\cite{sharma2015action}, VideoLSTM~\cite{li2018videolstm} and Actor Supervision~\cite{escorcia2020guess}. We also evaluate our approach against unsupervised action localization approaches such as Soomro \textit{et al.}~\cite{soomro2017unsupervised} and the closely related predictive learning approach~\cite{aakur2020action}, which we term as PredLearn. 
Note, we compare against PredLearn when the number of clusters is set to the ground-truth clusters ($k=k_{gt}$). 

\subsection{Quantitative Analysis}\label{sec:quant_results}
We first present the quantitative results of the proposed in Table~\ref{tab:sota}, where we compare against different baseline approaches. 
We report the mean average precision (mAP) scores over the most commonly reported overlap thresholds of $0.2$ and $0.5$. 
The approaches are ordered by the amount of supervision required for training. The models at the top require \textit{strong} supervision in terms of spatial annotations such as bounding boxes \textit{and} video-level labels to localize and classify the action. The models in the middle are \textit{weakly} supervised and hence only require video-level labels for training. The approaches at the bottom require no training annotations. It can be seen that our approach outperforms all baselines, including fully supervised models, on the \textit{UCF Sports} dataset, even at higher thresholds. Interestingly, we significantly outperform the closely related PredLearn by a significant margin ($\approx 15\%$ in absolute mAP). 

On datasets with significantly higher complexity, such as JHMDB and THUMOS'13, we see consistent improvements over the other unsupervised models such as PredLearn and Soomro \textit{et al.}'s action discovery approach, that use bounding box proposals from \textit{class-specific} proposals and hence are restricted to objects (humans) that are present in the pre-trained object detection models. On the other hand, we use \textit{class-agnostic} proposals and are not restricted to any object class. Also, it is interesting to note that hierarchical predictive learning and actor-centered feature representations help overcome the challenges posed by occlusions and clutter, as indicated by the significant gains over PredLearn at higher thresholds on JHMDB and THUMOS'13. 

\begin{table}[t]
    \centering
    {
    \begin{tabular}{|c|c|c|c|c|c|c|c|c|c|}
    \toprule
    \multirow{1}{*}{{Test Data }$\rightarrow$} & \multicolumn{2}{|c|}{{UCF Sports}} & \multicolumn{2}{|c|}{{JHMDB}} & \multicolumn{2}{|c|}{{THUMOS'13}}\\
    \cline{2-7}
     \multirow{2}{*}{{Train Data }$\downarrow$} & AC-HPL & PredLearn & AC-HPL & PredLearn & AC-HPL & PredLearn \\
     \cline{2-7}
     & \multicolumn{2}{|c|}{{$\sigma{=}$0.5}} & \multicolumn{2}{|c|}{{$\sigma{=}$0.2}} & \multicolumn{2}{|c|}{{$\sigma{=}$0.2}} \\
    \toprule
    {UCF Sports} & \textbf{0.59} & {0.32} & \textbf{0.39} & {0.19} & \textbf{0.38} & {0.20}\\\hline
    {JHMDB} & \textbf{0.48} & {0.23} & \textbf{0.43} & {0.30} & \textbf{0.35} & {0.26}\\\hline
    {THUMOS'13} & \textbf{0.50} & {0.27} & \textbf{0.40} & {0.24}) & \textbf{0.38} & {0.31} \\
    \bottomrule
    \end{tabular}
    }
    \setlength{\belowcaptionskip}{-15pt}
    \caption{Generalization capability when evaluated on out-of-domain test samples \textit{without finetuning}. PredLearn refers to ~\cite{aakur2020action} and AC-HPL refers to our approach.}
    \label{tab:loc_perf}
\end{table}

\begin{table*}[t]
    \centering
    \resizebox{0.99\columnwidth}{!}
    {
    \begin{tabular}{ cc }   
    \begin{tabular}{|c|c|c|c|}
        \toprule
        \multirow{2}{*}{\textbf{Approach}} & \multicolumn{2}{|c|}{\textbf{Supervision?}}  & \multirow{2}{*}{\textbf{Acc.}} \\
        \cline{2-3}
        & \textbf{Label} & \textbf{Box} & \\
        \midrule
        LRCN~\cite{qi2018stagnet} & \ding{51} & \ding{51} & 64.0 \\ 
        VGG-16~\cite{qi2018stagnet} & \ding{51} & \ding{55} & 68.3 \\
        VGG-16~\cite{qi2018stagnet} & \ding{51} & \ding{51} & 71.2 \\
        Hierarchical LSTM~\cite{ibrahim2016hierarchical} & \ding{51} & \ding{51} & 81.1 \\
        CERN~\cite{shu2017cern} & \ding{51} & \ding{51} & 84.8 \\
        stagNET~\cite{qi2018stagnet} & \ding{51} & \ding{51} & 89.1 \\
        ARG~\cite{wu2019learning} & \ding{51} & \ding{51} & 91.0 \\
        Action Transformer~\cite{gavrilyuk2020actor} & \ding{51} & \ding{51} & 92.8 \\
        GroupFormer~\cite{li2021groupformer} & \ding{51} & \ding{51} & 96.3 \\
        \midrule
        AC-HPL (k-means)$^*$ & \ding{55} & \ding{55} & 72.2 \\
        AC-HPL (Finetuned)$^*$ & \ding{51} & \ding{55} & \textbf{80.2} \\
        \bottomrule
    \end{tabular}
    & \quad
    \begin{tabular}{|c|c|c|c|c|}
        \toprule
        \multirow{2}{*}{\textbf{Approach}} & \multirow{2}{*}{\textbf{Avg. IOU}} & \multirow{2}{*}{\textbf{Recall}} & \multicolumn{2}{|c|}{\textbf{mAP}}\\
        \cline{4-5}
        & & & \textbf{0.2} & \textbf{0.5}\\
        \midrule
        PredLearn     &  0.18	& 0.172	& 0.378 & 0.011 \\ 
        AC-HPL (K=1)  &  0.205 & 0.181 & 0.442 & 0.017 \\
        AC-HPL (K=5)  &  0.271 & 0.284 & 0.545 & 0.068 \\
        AC-HPL (K=10) &  0.342 & 0.396 & 0.551 & 0.14 \\
        AC-HPL (K=25) &  0.472 & 0.634 & 0.723 & 0.449 \\
        \bottomrule
        \multicolumn{5}{c}{$^*$ denotes features are trained for only 1 epoch on the target dataset.}\\
        
    \end{tabular}\\
    (a) & (b)\\
    \end{tabular}
    }
    \setlength{\belowcaptionskip}{-15pt}
    \caption{Evaluation of our approach on the \textbf{Collective Activities} dataset for (a) multi-actor group activity recognition and (b) multi-actor group activity localization. }
    \label{tab:collective_results}
\end{table*}

\textbf{Generalization to Novel Domains.} In addition to evaluating the proposed approach in traditional settings, we also assess its ability to generalize to \textit{novel} domains. To be specific, we check its generalization capability by training on one dataset and testing its performance on a different dataset \textit{without finetuning}. 
We begin by evaluating the approach on the generalization task by training the model on the training data from one of the three standard benchmarks (UCF Sports, JHMDB - Split 1, and THUMOS'13) and evaluating on the others. While the three datasets have similar actions, they have varying amounts of data, camera motion, and occlusions, which provide a challenging benchmark for evaluating generalization performance. 
We also report the closely related PredLearn approach's performance, which does not use hierarchical prediction and actor-centered representations. Table~\ref{tab:loc_perf} summarizes the results. It can be seen that the proposed approach generalizes well across datasets, \textit{regardless of the training data size}. 
For example, UCF Sports has a very small number of training data ($103$) and classes ($10$). However, the model can transfer well to other datasets with more classes (21 for JHMDB and 24 for THUMOS'13). 
It is to be noted that the model is \textit{not finetuned} on any data in the target domain yet performs as well as weakly supervised models such as VideoLSTM ($0.37$ mAP@0.2 on THUMOS'13), \textit{which was trained on the data.} 
Similarly, the use of actor-centered representations allows for better generalization compared to PredLearn, which has a poorer recognition performance due to the lack of \textit{contextualized} feature representations. 

\subsection{Multi-Actor Group Activity Localization}
Our approach can be naturally extended to multi-actor group activity recognition and localization using the Collective Activities dataset~\cite{choi2009they}. The goal is to recognize the collective or group activity performed by the majority of the actors in the scene. While majority of the prior works~\cite{qi2018stagnet,shu2017cern,wu2019learning,gavrilyuk2020actor,li2021groupformer} have focused on \textit{recognition}, there have not been efforts for multi-actor \textit{localization}. 
Using the prediction errors outlined in Section~\ref{sec:localization}, we can increase the number of attention \textit{grids} ($K_{attn}$) and use the resulting attention points to localize multiple actors in a given scene. 
We show that our approach can attend to multiple actors and learn robust representations for simultaneous localization and recognition in multi-actor videos.
Table~\ref{tab:collective_results}(a) summarizes the performance for \textit{recognition}, while Table~\ref{tab:collective_results}(b) summarizes the performance for \textit{localization}. For recognition, we evaluate two versions of the proposed approach to generate the labels - a completely unsupervised version with k-means for prediction and a fine-tuned version where the features are categorized into classes using a 2-layer feedforward neural network. 
As can be seen, without extensive training and annotations such as bounding boxes, we can achieve a recognition of $80.2\%$, which is remarkable considering that other state-of-the-art approaches require large amounts of training annotations and epochs. We only need one epoch of training to finetune the features to the multi-actor setting and do not need any annotations for learning. 
For localization, we report the average IOU of all bounding boxes produced by the approach and the mAP at 0.2 IOU and 0.5 IOU. To evaluate the upper bound of the approach, we also compute recall by considering groundtruth bounding boxes with at least one attention point as a true positive (TP). As can be seen from Table~\ref{tab:collective_results}(b), as the number of attention points ($K_{attn}$) increases, we achieve a better IOU, recall, and mAP. Note that the closely related PredLearn is not able to handle multiple actor localization since their attention focuses only on the dominant \textit{actor} and considers other actors as clutter. Nevertheless, these results and the qualitative visualizations in Figure~\ref{fig:qualitative_examples} show that the approach can perform multi-actor localization without any bells and whistles while not being explicitly trained for the task. 

\subsection{Ablation Studies}
\begin{figure*}[t]
    \centering
    \begin{tabular}{ccc}
         \includegraphics[width=0.3\columnwidth]{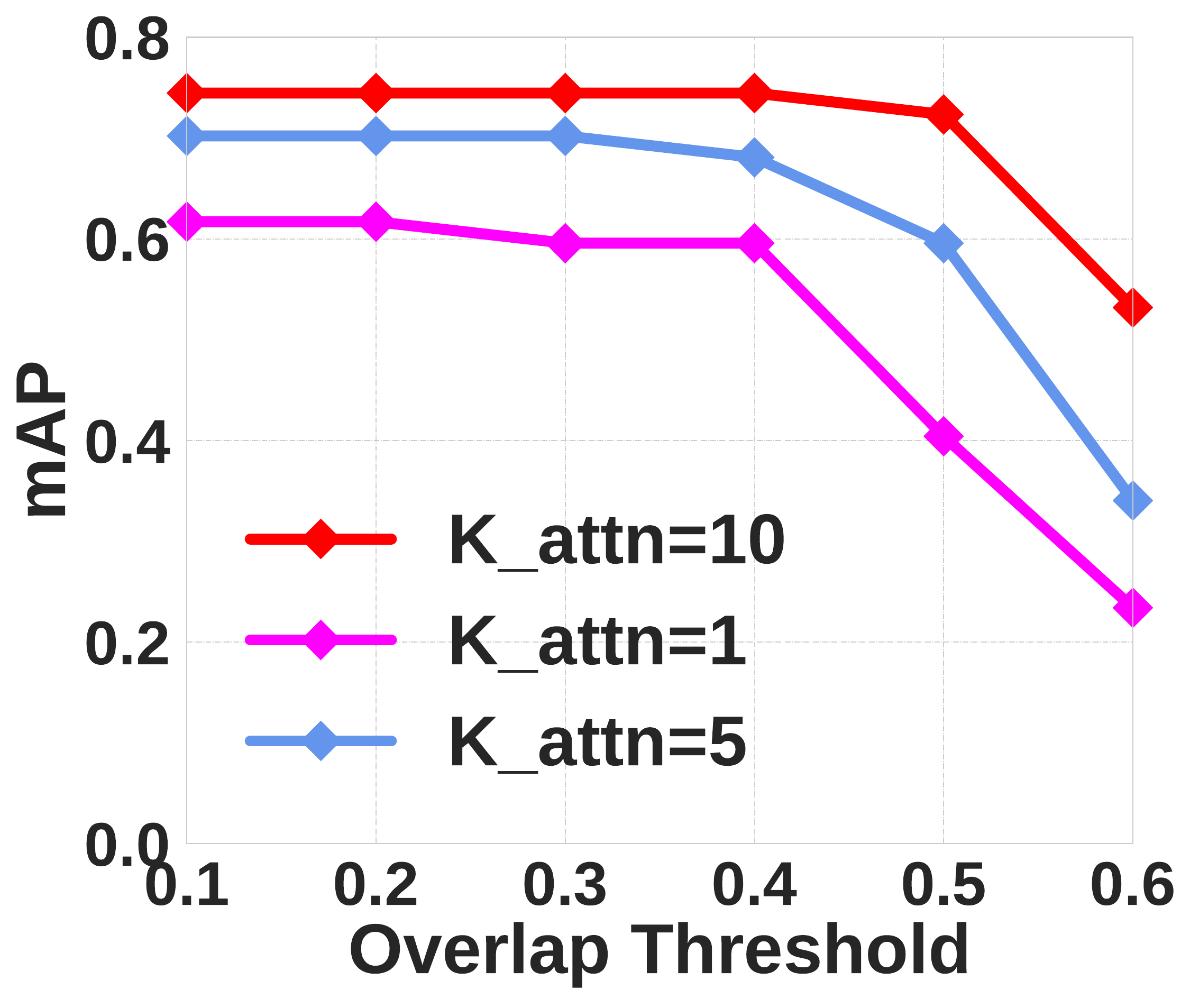} & \includegraphics[width=0.3\columnwidth]{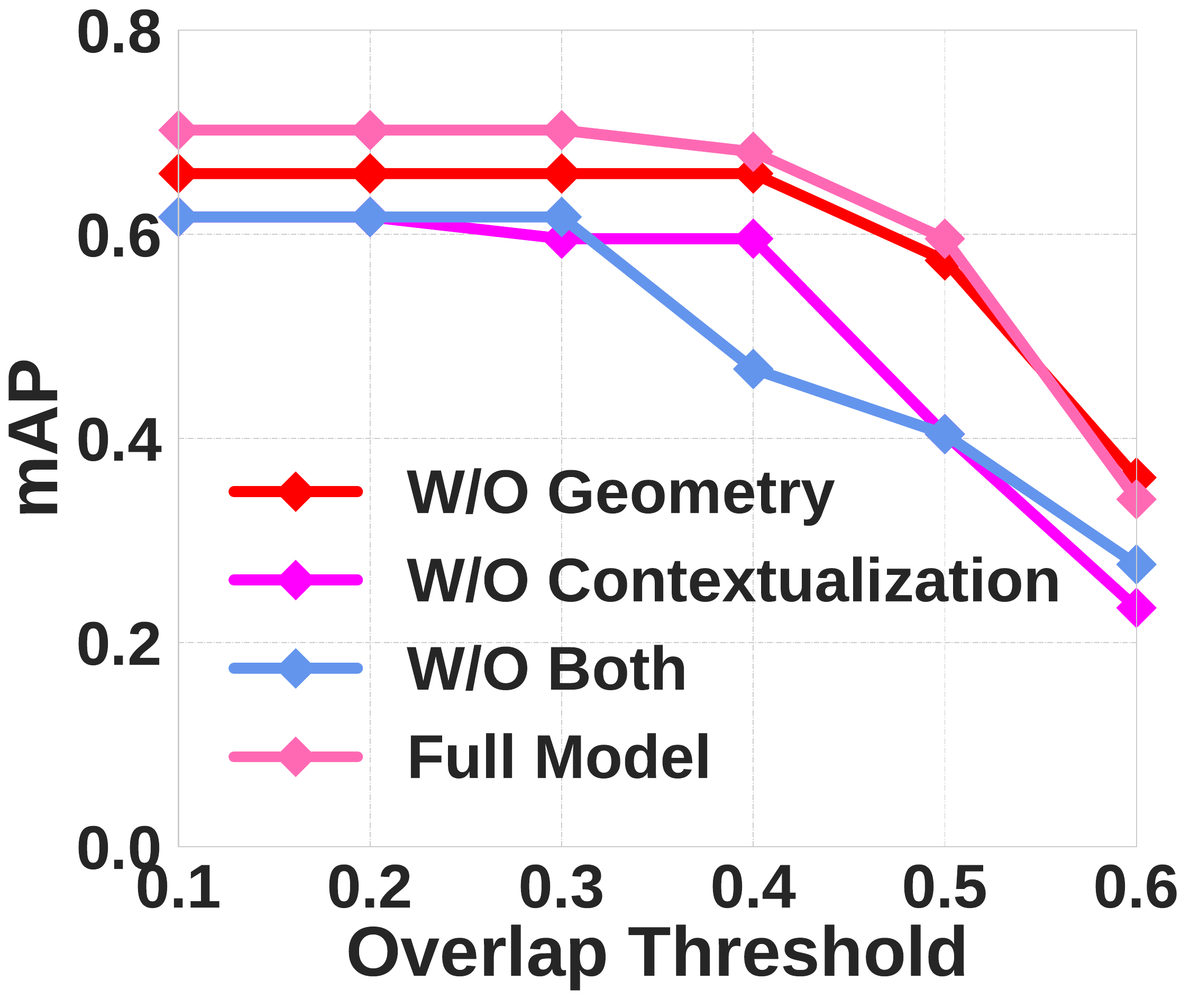} &
         \includegraphics[width=0.3\columnwidth]{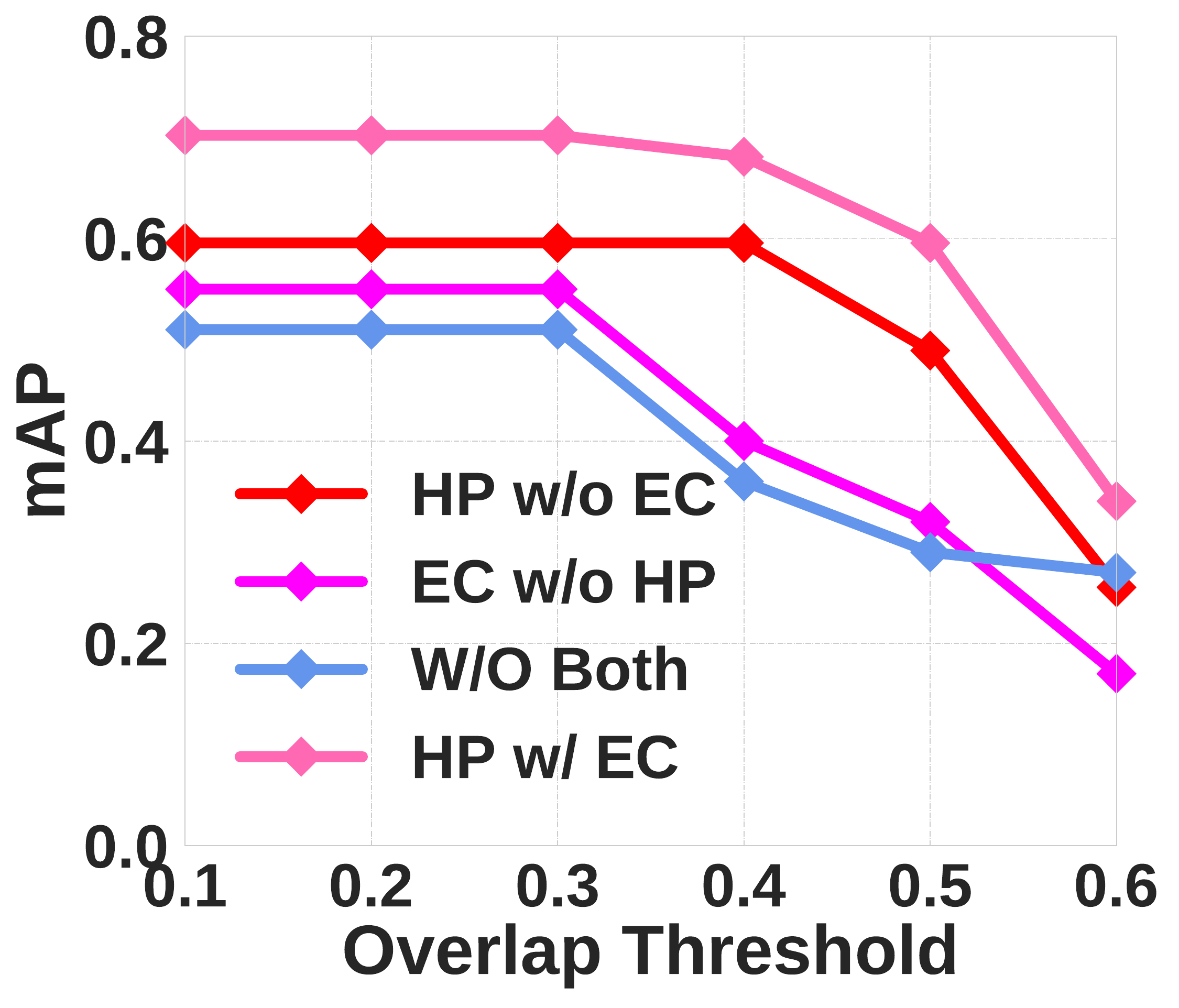} \\
         (a) & (b) & (c)\\
         \includegraphics[width=0.3\columnwidth]{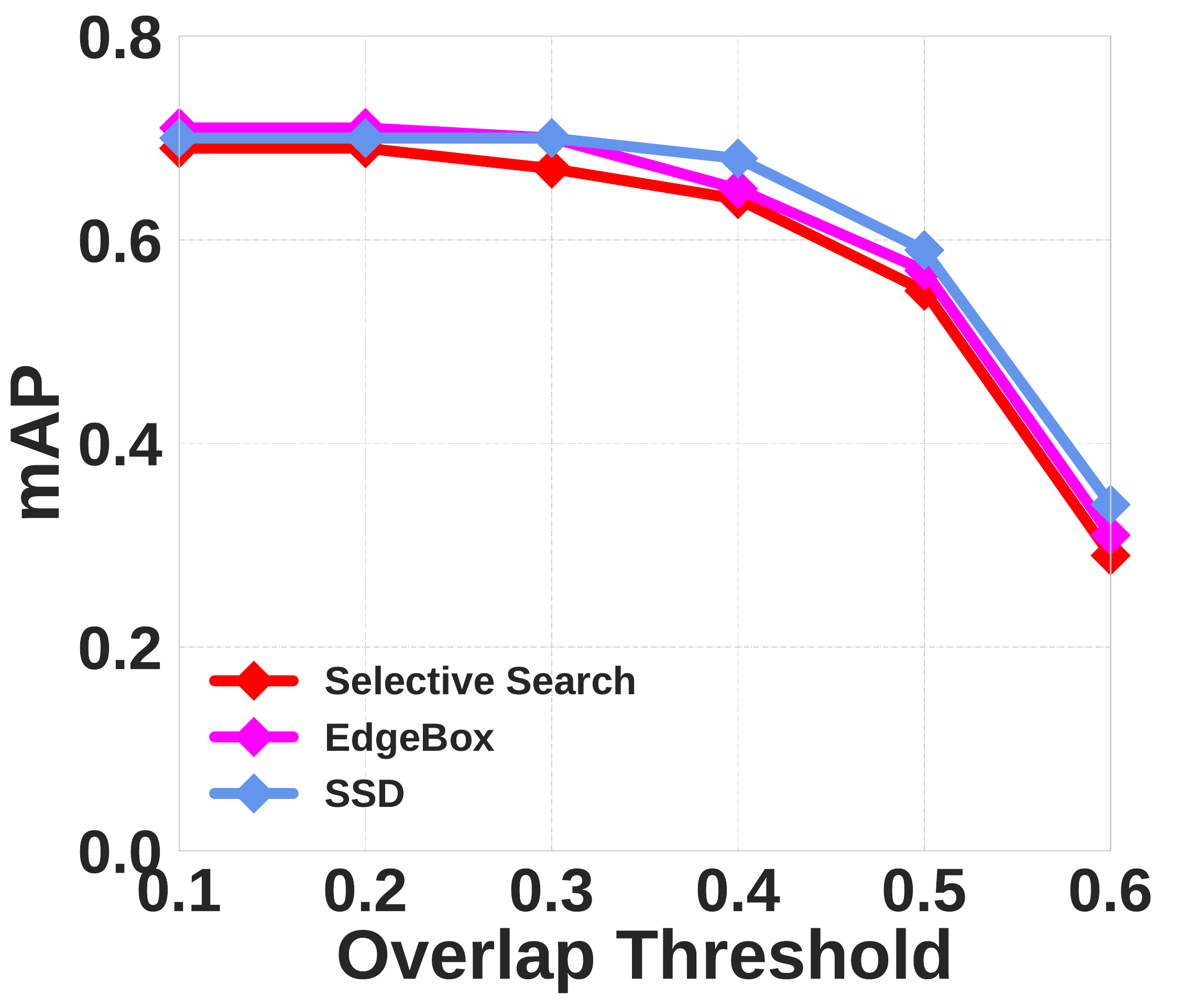} & 
         \includegraphics[width=0.3\columnwidth]{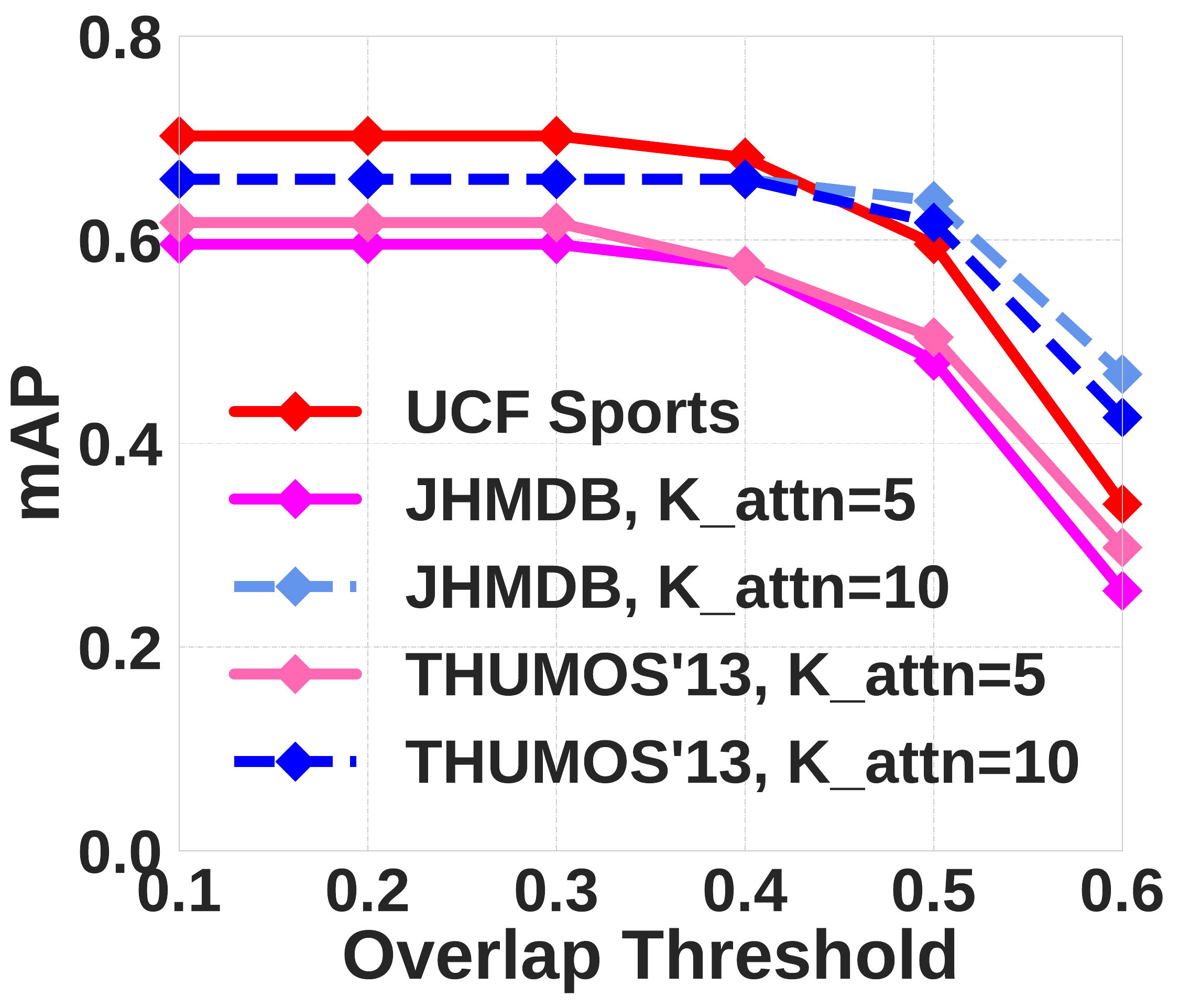} & 
         \includegraphics[width=0.3\columnwidth]{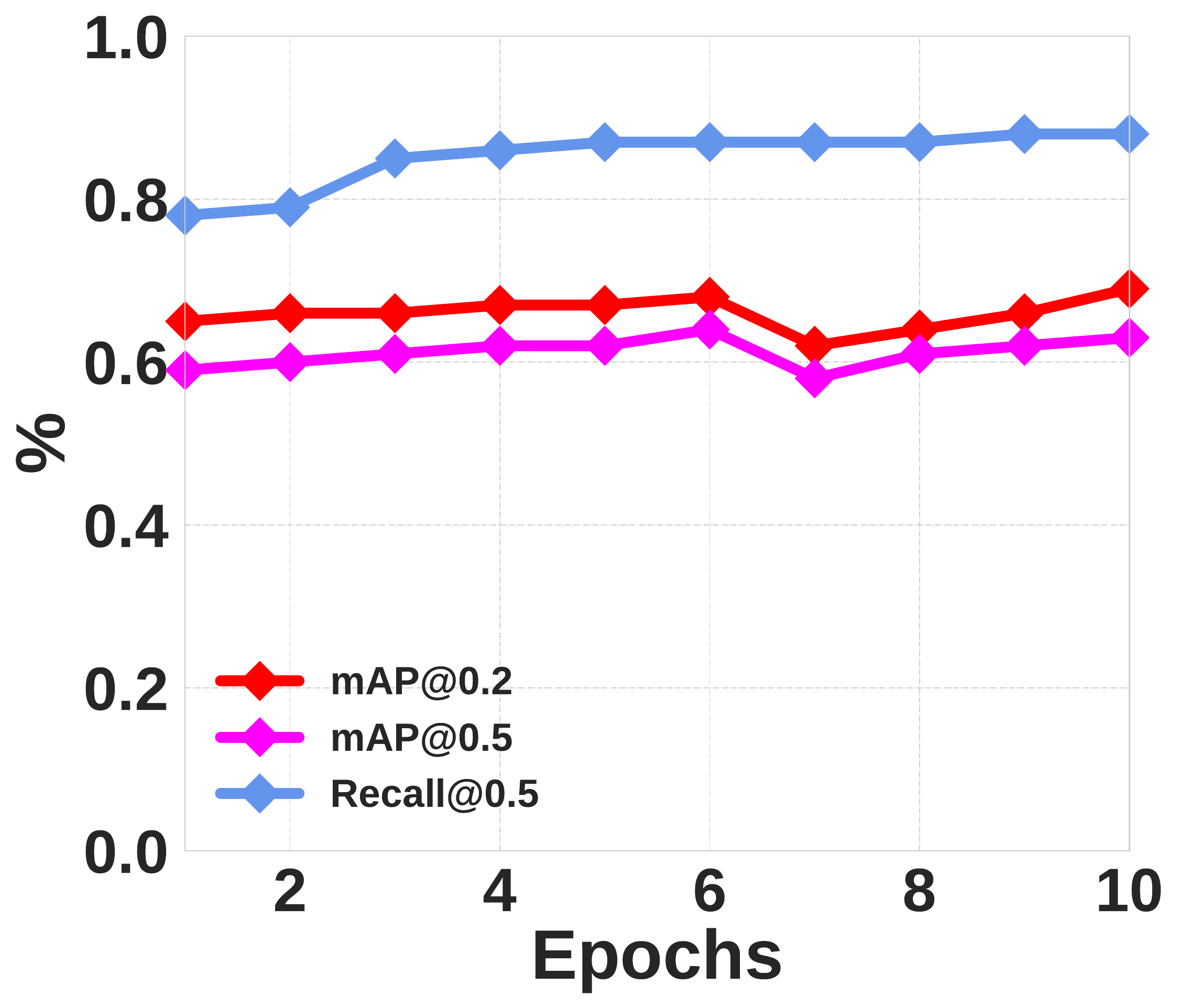} \\
         (d) & (e) & (f) \\

    \end{tabular}
    \setlength{\belowcaptionskip}{-15pt}
    \caption{Ablation experiments on UCF Sports to evaluate the effect of (a) number of attention ``grids'', (b) actor-centered prediction, and (c) event-centric perception, (d) choice of region proposal, (e) out-of-domain data, and (e) number of training epochs.}
    
    \label{fig:ablation_plots}
\end{figure*}

\begin{figure*}[t]
\centering
\begin{tabular}{c|c}
\toprule
\textbf{AC-HPL (Ours)} & \textbf{PredLearn}\\
\toprule
\addtolength{\tabcolsep}{-5pt}  
     \begin{tabular}{cccc}
       \includegraphics[width=0.113\textwidth]{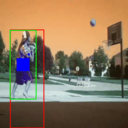} &  
       \includegraphics[width=0.113\textwidth]{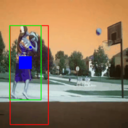} &  
       \includegraphics[width=0.113\textwidth]{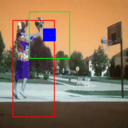} &  
       \includegraphics[width=0.113\textwidth]{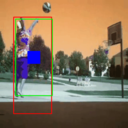} 
    \end{tabular} &  
    \addtolength{\tabcolsep}{-5pt}  
    \begin{tabular}{cccc}
         \includegraphics[width=0.113\textwidth]{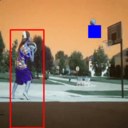}&
       \includegraphics[width=0.113\textwidth]{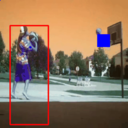} & 
       \includegraphics[width=0.113\textwidth]{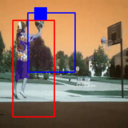} &
       \includegraphics[width=0.113\textwidth]{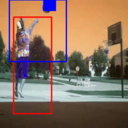}
    \end{tabular}\\

\addtolength{\tabcolsep}{-5pt}  
     \begin{tabular}{cccc}
       \includegraphics[width=0.113\textwidth]{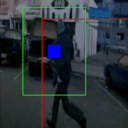}&
       \includegraphics[width=0.113\textwidth]{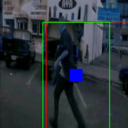} & 
       \includegraphics[width=0.113\textwidth]{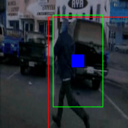} &
       \includegraphics[width=0.113\textwidth]{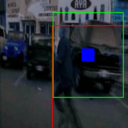}
    \end{tabular} &  
    \addtolength{\tabcolsep}{-5pt}  
    \begin{tabular}{cccc}
         \includegraphics[width=0.113\textwidth]{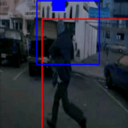} &  
       \includegraphics[width=0.113\textwidth]{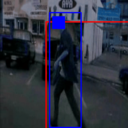} &  
       \includegraphics[width=0.113\textwidth]{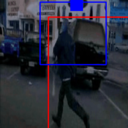} &  
       \includegraphics[width=0.113\textwidth]{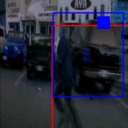} 
    \end{tabular}\\

\addtolength{\tabcolsep}{-5pt}  
     \begin{tabular}{cccc}
       \includegraphics[width=0.113\textwidth]{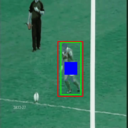} &  
       \includegraphics[width=0.113\textwidth]{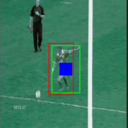} &  
       \includegraphics[width=0.113\textwidth]{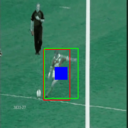} &  
       \includegraphics[width=0.113\textwidth]{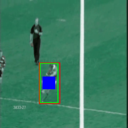} 
    \end{tabular} &  
    \addtolength{\tabcolsep}{-5pt}  
    \begin{tabular}{cccc}
         \includegraphics[width=0.113\textwidth]{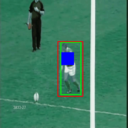}&
       \includegraphics[width=0.113\textwidth]{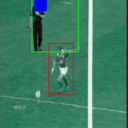} & 
       \includegraphics[width=0.113\textwidth]{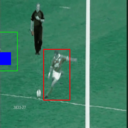} &
       \includegraphics[width=0.113\textwidth]{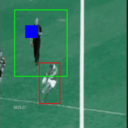}
    \end{tabular}\\
    \midrule
    \multicolumn{2}{c}{\textbf{Multi-Actor Localization}}\\
    \midrule
    \addtolength{\tabcolsep}{-5pt}  
     \begin{tabular}{cccc}
       \includegraphics[width=0.113\textwidth]{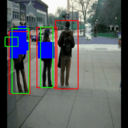} &  
       \includegraphics[width=0.113\textwidth]{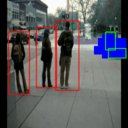} &  
       \includegraphics[width=0.113\textwidth]{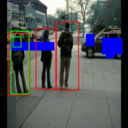} &  
       \includegraphics[width=0.113\textwidth]{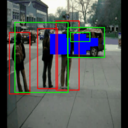} 
    \end{tabular} &  
    \addtolength{\tabcolsep}{-5pt}  
    \begin{tabular}{cccc}
         \includegraphics[width=0.113\textwidth]{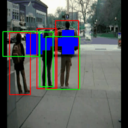} &  
       \includegraphics[width=0.113\textwidth]{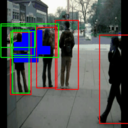} &  
       \includegraphics[width=0.113\textwidth]{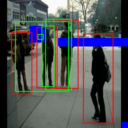} &  
       \includegraphics[width=0.113\textwidth]{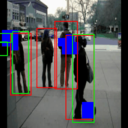} 
    \end{tabular}\\
    \addtolength{\tabcolsep}{-5pt}  
     \begin{tabular}{cccc}
       \includegraphics[width=0.113\textwidth]{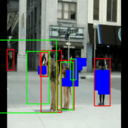} &  
       \includegraphics[width=0.113\textwidth]{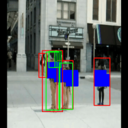} &  
       \includegraphics[width=0.113\textwidth]{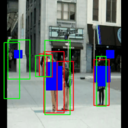} &  
       \includegraphics[width=0.113\textwidth]{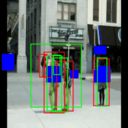} 
    \end{tabular} &  
    \addtolength{\tabcolsep}{-5pt}  
    \begin{tabular}{cccc}
         \includegraphics[width=0.113\textwidth]{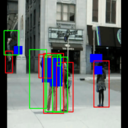}&
       \includegraphics[width=0.113\textwidth]{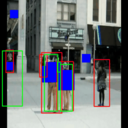} & 
       \includegraphics[width=0.113\textwidth]{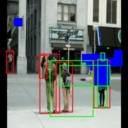} &
       \includegraphics[width=0.113\textwidth]{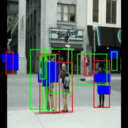}
    \end{tabular}\\
    \midrule
    \multicolumn{2}{c}{\textbf{Unsuccessful Localization}}\\
    \midrule
    \addtolength{\tabcolsep}{-5pt}  
     \begin{tabular}{cccc}
       \includegraphics[width=0.113\textwidth]{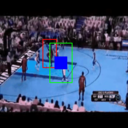} &  
       \includegraphics[width=0.113\textwidth]{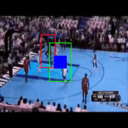} &  
       \includegraphics[width=0.113\textwidth]{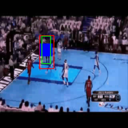} &  
       \includegraphics[width=0.113\textwidth]{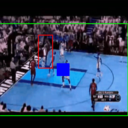} 
    \end{tabular} &  
    \addtolength{\tabcolsep}{-5pt}  
    \begin{tabular}{cccc}
         \includegraphics[width=0.113\textwidth]{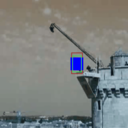}&
       \includegraphics[width=0.113\textwidth]{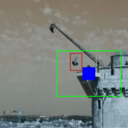} & 
       \includegraphics[width=0.113\textwidth]{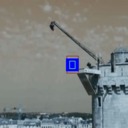} &
       \includegraphics[width=0.113\textwidth]{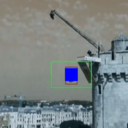}
    \end{tabular}\\
    \bottomrule
\end{tabular}
    \setlength{\belowcaptionskip}{-15pt}
    \caption{\textbf{Qualitative Examples.} \textit{Top:} handling camera motion and background motion to maintain context in localization. \textit{Middle}: multi-actor localizations from AC-HPL. \textit{Bottom}: Unsuccessful localizations from AC-HPL. \textit{Visualization Legend:} Red BB: Grountruth, Green BB: Predictions, Blue Squares: Attention Locations.
    }
    \label{fig:qualitative_examples}
\end{figure*}

In Figure~\ref{fig:ablation_plots}(a) we show the effect of the changing the number of attention ``\textit{grids}'' ($K_{attn}$) (Section~\ref{sec:localization}). 
As the $K_{attn}$ increases, the localization performance also increases and allows the model to keep track of the object of interest even if there are other potential actors. 
We also evaluate the effect of the different terms in the actor-centered prediction loss (Equation~\ref{eqn:objectPred}). As can be seen from Figure~\ref{fig:qualitative_examples}(b), the use of both geometry prediction and contextualized feature prediction help improve the performance significantly, with the use of contextualized prediction providing a greater jump in performance. In Figure~\ref{fig:qualitative_examples}(c), we present the effect of using event-centric perceptual features (Section~\ref{sec:event-centric-percep}) on the framework with and without hierarchical prediction. It can be that the use of both improves the performance, especially at higher thresholds, indicating that the use of hierarchical prediction with event-centric features helps attend to areas of interest.

\textit{Effect of Region Proposal Methods.} To evaluate the effect of object detection modules on the proposed approach, we try other bounding box proposals from \textit{untrained} approaches such as EdgeBox and Selective Search. We present the results on UCF Sports below in Figure~\ref{fig:qualitative_examples}(d). As can be seen, our approach is not dependent on SSD as the object proposal mechanism and is able to use any region proposal mechanism as its input. 
We do not use any class labels from SSD and make it class-agnostic to ensure that we do not have any assumptions about the actor or domain semantics, unlike supervised or weakly-supervised approaches that use object detectors as part of the action proposals. 

\textit{Effect of out-of-domain data.} We also evaluated the localization performance of the approach when trained with out-of-domain data. Figure~\ref{fig:ablation_plots}(e) shows that the real performance drop is at higher overlap thresholds when testing on data outside of the training domain. However, increasing $K_{attn}$ helps alleviate this issue and even outperforms models trained in the same domain. 

\textit{Effect of multiple training epochs.} Although our approach is designed to work with one epoch of training, we also evaluate the impact of multi-epoch training on the UCF Sports dataset and present the results in Figure~\ref{fig:ablation_plots}(f). It can be seen that increasing the number of epochs allows the model to learn better features for recognition while the localization is improved as well. 

\subsection{Qualitative Analysis}\label{sec:qualititative}
We qualitatively analyze our approach and visualize some interesting instances in Figure~\ref{fig:qualitative_examples}. We show two specific groups of examples - (i) a comparison with PredLearn~\cite{aakur2020action} to highlight the importance of using actor-centered features and hierarchical prediction beyond numbers presented in Section~\ref{sec:results}, and (ii) some failure modes of the approach to identify possible ways to mitigate them. 
In rows 1 and 4, it can be seen that although there are other objects in the scene whose motion is unpredictable, the use of multiple attention grids and actor-centered prediction helps the model to maintain focus on the actor. Row 3 shows that the model can overcome the challenges posed by camera motion \textit{and} object deformation to maintain context in prediction, whereas PredLearn (without hierarchical prediction) is influenced by the camera motion and loses track of the object. 
We also visualize some of the failure modes of the proposed model in the final row in Figure~\ref{fig:qualitative_examples}. In particular, we would like to highlight two areas that lead to failure. First, consider the sequence on the left. The model's attention is initially on the wrong player in the scene and continues to attend to areas of the same player, which we attribute to the actor-centered prediction. 
Although the object is well localized, it is not the \textit{labeled} object of interest to which the model does not have access. The second failure mode, highlighted in the sequence on the right, is bounding box selection. Although the attention is on the object for most frames, \textit{class-agnostic} proposals returns poorer bounding box fit. 
The mAP score does not take these factors into account. 

\section{Conclusion}\label{sec:conclusion}
We showed that we can learn actor-centered representations to localize actions with just a single-pass through video for training (single epoch) and without training labels and outlines. We do not need multiple training epochs to build the representations. This makes the approach useful for many real-world context where storing the video raises privacy or high storage cost concerns. Our solution was a hierarchical predictive learning framework that continuously predicts and learns from errors at different granularities. 
The resulting spatial-temporal error localized the action. The model leverages a novel actor-centered representation to learn robust features that mitigate the effect of camera motion and background clutter. We showed that we can beat SOTA on unsupervised action localization and multi-actor group activity localization while generalizing to novel domains \textit{without finetuning}. 

\textbf{Acknowledgements.} This research was supported in part by the US National Science Foundation grants CNS 1513126, IIS 1956050, IIS 2143150, and IIS 1955230.
%
%
\bibliographystyle{splncs04}
\bibliography{egbib}

\begin{thebibliography}{10}
\providecommand{\url}[1]{\texttt{#1}}
\providecommand{\urlprefix}{URL }
\providecommand{\doi}[1]{https://doi.org/#1}

\bibitem{aakur2019wacv}
Aakur, S., de~Souza, F.D., Sarkar, S.: Going deeper with semantics: Exploiting
  semantic contextualization for interpretation of human activity in videos.
  In: IEEE Winter Conference on Applications of Computer Vision (WACV). IEEE
  (2019)

\bibitem{Aakur_2019_CVPR}
Aakur, S.N., Sarkar, S.: A perceptual prediction framework for self supervised
  event segmentation. In: The IEEE Conference on Computer Vision and Pattern
  Recognition (CVPR) (June 2019)

\bibitem{aakur2020action}
Aakur, S.N., Sarkar, S.: Action localization through continual predictive
  learning. arXiv preprint arXiv:2003.12185  (2020)

\bibitem{bahdanau2014neural}
Bahdanau, D., Cho, K., Bengio, Y.: Neural machine translation by jointly
  learning to align and translate. arXiv preprint arXiv:1409.0473  (2014)

\bibitem{choi2009they}
Choi, W., Shahid, K., Savarese, S.: What are they doing?: Collective activity
  classification using spatio-temporal relationship among people. In: 2009 IEEE
  12th International Conference on Computer Vision Workshops. pp. 1282--1289.
  IEEE (2009)

\bibitem{escorcia2020guess}
Escorcia, V., Dao, C.D., Jain, M., Ghanem, B., Snoek, C.: Guess where?
  actor-supervision for spatiotemporal action localization. Computer Vision and
  Image Understanding  \textbf{192},  102886 (2020)

\bibitem{gan2018geometry}
Gan, C., Gong, B., Liu, K., Su, H., Guibas, L.J.: Geometry guided convolutional
  neural networks for self-supervised video representation learning. In:
  Proceedings of the IEEE Conference on Computer Vision and Pattern
  Recognition. pp. 5589--5597 (2018)

\bibitem{gavrilyuk2020actor}
Gavrilyuk, K., Sanford, R., Javan, M., Snoek, C.G.: Actor-transformers for
  group activity recognition. In: Proceedings of the IEEE/CVF Conference on
  Computer Vision and Pattern Recognition. pp. 839--848 (2020)

\bibitem{gkioxari2015finding}
Gkioxari, G., Malik, J.: Finding action tubes. In: Proceedings of the IEEE
  Conference on Computer Vision and Pattern Recognition. pp. 759--768 (2015)

\bibitem{hochreiter1997long}
Hochreiter, S., Schmidhuber, J.: Long short-term memory. Neural Computation
  \textbf{9}(8),  1735--1780 (1997)

\bibitem{horstmann2015surprise}
Horstmann, G., Herwig, A.: Surprise attracts the eyes and binds the gaze.
  Psychonomic Bulletin \& Review  \textbf{22}(3),  743--749 (2015)

\bibitem{hou2017tube}
Hou, R., Chen, C., Shah, M.: Tube convolutional neural network (t-cnn) for
  action detection in videos. In: Proceedings of the IEEE International
  Conference on Computer Vision (ICCV). pp. 5822--5831 (2017)

\bibitem{ibrahim2016hierarchical}
Ibrahim, M.S., Muralidharan, S., Deng, Z., Vahdat, A., Mori, G.: A hierarchical
  deep temporal model for group activity recognition. In: Proceedings of the
  IEEE Conference on Computer Vision and Pattern Recognition. pp. 1971--1980
  (2016)

\bibitem{jain2017tubelets}
Jain, M., Van~Gemert, J., J{\'e}gou, H., Bouthemy, P., Snoek, C.G.: Tubelets:
  Unsupervised action proposals from spatiotemporal super-voxels. International
  Journal of Computer Vision  \textbf{124}(3),  287--311 (2017)

\bibitem{jhuang2013towards}
Jhuang, H., Gall, J., Zuffi, S., Schmid, C., Black, M.J.: Towards understanding
  action recognition. In: Proceedings of the IEEE International Conference on
  Computer Vision. pp. 3192--3199 (2013)

\bibitem{ji2019invariant}
Ji, X., Henriques, J.F., Vedaldi, A.: Invariant information clustering for
  unsupervised image classification and segmentation. In: Proceedings of the
  IEEE International Conference on Computer Vision. pp. 9865--9874 (2019)

\bibitem{jiang2014thumos}
Jiang, Y.G., Liu, J., Zamir, A.R., Toderici, G., Laptev, I., Shah, M.,
  Sukthankar, R.: Thumos challenge: Action recognition with a large number of
  classes (2014)

\bibitem{kuehne2014language}
Kuehne, H., Arslan, A., Serre, T.: The language of actions: Recovering the
  syntax and semantics of goal-directed human activities. In: IEEE Conference
  on Computer Vision and Pattern Recognition (CVPR). pp. 780--787 (2014)

\bibitem{lan2011discriminative}
Lan, T., Wang, Y., Mori, G.: Discriminative figure-centric models for joint
  action localization and recognition. In: 2011 International Conference on
  Computer Vision. pp. 2003--2010. IEEE (2011)

\bibitem{li2021groupformer}
Li, S., Cao, Q., Liu, L., Yang, K., Liu, S., Hou, J., Yi, S.: Groupformer:
  Group activity recognition with clustered spatial-temporal transformer. In:
  Proceedings of the IEEE/CVF International Conference on Computer Vision. pp.
  13668--13677 (2021)

\bibitem{li2018videolstm}
Li, Z., Gavrilyuk, K., Gavves, E., Jain, M., Snoek, C.G.: Videolstm convolves,
  attends and flows for action recognition. Computer Vision and Image
  Understanding  \textbf{166},  41--50 (2018)

\bibitem{lin2013network}
Lin, M., Chen, Q., Yan, S.: Network in network. arXiv preprint arXiv:1312.4400
  (2013)

\bibitem{liu2016ssd}
Liu, W., Anguelov, D., Erhan, D., Szegedy, C., Reed, S., Fu, C.Y., Berg, A.C.:
  Ssd: Single shot multibox detector. In: European Conference on Computer
  Vision. pp. 21--37. Springer (2016)

\bibitem{liu2020real}
Liu, Y., Tu, Z., Lin, L., Xie, X., Qin, Q.: Real-time spatio-temporal action
  localization via learning motion representation. In: Proceedings of the Asian
  Conference on Computer Vision (2020)

\bibitem{luong2015effective}
Luong, M.T., Pham, H., Manning, C.D.: Effective approaches to attention-based
  neural machine translation. arXiv preprint arXiv:1508.04025  (2015)

\bibitem{pan2021actor}
Pan, J., Chen, S., Shou, M.Z., Liu, Y., Shao, J., Li, H.: Actor-context-actor
  relation network for spatio-temporal action localization. In: Proceedings of
  the IEEE/CVF Conference on Computer Vision and Pattern Recognition. pp.
  464--474 (2021)

\bibitem{pramono2019hierarchical}
Pramono, R.R.A., Chen, Y.T., Fang, W.H.: Hierarchical self-attention network
  for action localization in videos. In: Proceedings of the IEEE/CVF
  International Conference on Computer Vision. pp. 61--70 (2019)

\bibitem{qi2018stagnet}
Qi, M., Qin, J., Li, A., Wang, Y., Luo, J., Van~Gool, L.: stagnet: An attentive
  semantic rnn for group activity recognition. In: Proceedings of the European
  Conference on Computer Vision (ECCV). pp. 101--117 (2018)

\bibitem{redmon2017yolo9000}
Redmon, J., Farhadi, A.: Yolo9000: better, faster, stronger. In: Proceedings of
  the IEEE Conference on Computer Vision and Pattern Recognition. pp.
  7263--7271 (2017)

\bibitem{rodriguez2008action}
Rodriguez, M.D., Ahmed, J., Shah, M.: Action mach a spatio-temporal maximum
  average correlation height filter for action recognition. In: 2008 IEEE
  Conference on Computer Vision and Pattern Recognition. pp.~1--8. IEEE (2008)

\bibitem{ILSVRC15}
Russakovsky, O., Deng, J., Su, H., Krause, J., Satheesh, S., Ma, S., Huang, Z.,
  Karpathy, A., Khosla, A., Bernstein, M., Berg, A.C., Fei-Fei, L.: {ImageNet
  Large Scale Visual Recognition Challenge}. International Journal of Computer
  Vision (IJCV)  \textbf{115}(3),  211--252 (2015).
  \doi{10.1007/s11263-015-0816-y}

\bibitem{sharma2015action}
Sharma, S., Kiros, R., Salakhutdinov, R.: Action recognition using visual
  attention. In: Neural Information Processing Systems: Time Series Workshop
  (2015)

\bibitem{shu2017cern}
Shu, T., Todorovic, S., Zhu, S.C.: Cern: confidence-energy recurrent network
  for group activity recognition. In: Proceedings of the IEEE Conference on
  Computer Vision and Pattern Recognition. pp. 5523--5531 (2017)

\bibitem{simonyan2014very}
Simonyan, K., Zisserman, A.: Very deep convolutional networks for large-scale
  image recognition. arXiv preprint arXiv:1409.1556  (2014)

\bibitem{soomro2015action}
Soomro, K., Idrees, H., Shah, M.: Action localization in videos through context
  walk. In: Proceedings of the IEEE International Conference on Computer
  Vision. pp. 3280--3288 (2015)

\bibitem{soomro2016predicting}
Soomro, K., Idrees, H., Shah, M.: Predicting the where and what of actors and
  actions through online action localization. In: Proceedings of the IEEE
  Conference on Computer Vision and Pattern Recognition. pp. 2648--2657 (2016)

\bibitem{soomro2017unsupervised}
Soomro, K., Shah, M.: Unsupervised action discovery and localization in videos.
  In: Proceedings of the IEEE International Conference on Computer Vision. pp.
  696--705 (2017)

\bibitem{soomro2012ucf101}
Soomro, K., Zamir, A.R., Shah, M.: Ucf101: A dataset of 101 human actions
  classes from videos in the wild. arXiv preprint arXiv:1212.0402  (2012)

\bibitem{tian2013spatiotemporal}
Tian, Y., Sukthankar, R., Shah, M.: Spatiotemporal deformable part models for
  action detection. In: Proceedings of the IEEE Conference on Computer Vision
  and Pattern Recognition. pp. 2642--2649 (2013)

\bibitem{tran2015learning}
Tran, D., Bourdev, L., Fergus, R., Torresani, L., Paluri, M.: Learning
  spatiotemporal features with 3d convolutional networks. In: Proceedings of
  the IEEE International Conference on Computer Vision. pp. 4489--4497 (2015)

\bibitem{tran2012max}
Tran, D., Yuan, J.: Max-margin structured output regression for spatio-temporal
  action localization. In: Advances in neural information processing systems.
  pp. 350--358 (2012)

\bibitem{wang2019self}
Wang, J., Jiao, J., Bao, L., He, S., Liu, Y., Liu, W.: Self-supervised
  spatio-temporal representation learning for videos by predicting motion and
  appearance statistics. In: Proceedings of the IEEE Conference on Computer
  Vision and Pattern Recognition. pp. 4006--4015 (2019)

\bibitem{wang2014video}
Wang, L., Qiao, Y., Tang, X.: Video action detection with relational
  dynamic-poselets. In: European Conference on Computer Vision. pp. 565--580.
  Springer (2014)

\bibitem{wang2017recurrent}
Wang, M., Ni, B., Yang, X.: Recurrent modeling of interaction context for
  collective activity recognition. In: Proceedings of the IEEE Conference on
  Computer Vision and Pattern Recognition. pp. 3048--3056 (2017)

\bibitem{weinzaepfel2015learning}
Weinzaepfel, P., Harchaoui, Z., Schmid, C.: Learning to track for
  spatio-temporal action localization. In: Proceedings of the IEEE
  international conference on computer vision. pp. 3164--3172 (2015)

\bibitem{wu2019learning}
Wu, J., Wang, L., Wang, L., Guo, J., Wu, G.: Learning actor relation graphs for
  group activity recognition. In: Proceedings of the IEEE/CVF Conference on
  Computer Vision and Pattern Recognition. pp. 9964--9974 (2019)

\bibitem{xie2016unsupervised}
Xie, J., Girshick, R., Farhadi, A.: Unsupervised deep embedding for clustering
  analysis. In: International Conference on Machine Learning (ICML). pp.
  478--487 (2016)

\bibitem{zhang2020learning}
Zhang, D., He, L., Tu, Z., Zhang, S., Han, F., Yang, B.: Learning motion
  representation for real-time spatio-temporal action localization. Pattern
  Recognition  \textbf{103},  107312 (2020)

\end{thebibliography}
\end{document}